\documentclass{article}
\usepackage{lineno}
\usepackage[colorlinks,bookmarksopen,bookmarksnumbered,citecolor=blue,urlcolor=red]{hyperref}
\usepackage{xcolor}
\usepackage{mathtools}
\usepackage{amsthm}
\usepackage{amsmath}
\usepackage{graphicx}
\usepackage[algo2e,linesnumbered,ruled,vlined]{algorithm2e}
\usepackage[noend]{algpseudocode}
\usepackage{multirow,tabularx}
\usepackage[utf8]{inputenc}
\usepackage{comment}  
\usepackage{amsfonts} 
\usepackage{booktabs}
\theoremstyle{definition}
\newtheorem{definition}{Definition}
\usepackage{authblk}

\newcommand{\ra}[1]{\renewcommand{\arraystretch}{#1}}
\modulolinenumbers[5]

\title{Predicting vehicles parking behaviour in shared premises for aggregated EV electricity demand response programs}

\begin{document}

\author{Vinicius Monteiro de Lira \thanks{Institute of Information Science and Technologies, Italian National Research Council, Pisa, Italy, viniciusmonteirodelira@isti.cnr.it}, Fabiano Pallonetto\thanks{School of Business, Maynooth University, Kildare, Ireland, fabiano.pallonetto@mu.ie}, Lorenzo Gabrielli\thanks{Institute of Information Science and Technologies, Italian National Research Council, Pisa, Italy, lorenzo.gabrielli@isti.cnr.it}, Chiara Renso \thanks{Institute of Information Science and Technologies, Italian National Research Council, Pisa, Italy, chiara.renso@isti.cnr.it}}

\date{}


\maketitle

\begin{abstract}
The global electric car sales in 2020 continued to exceed the expectations climbing to over 3 millions and reaching a market share of over 4\%. However, uncertainty of generation caused by higher penetration of renewable energies and the advent of Electrical Vehicles (EV) with their additional electricity demand could cause strains to the power system, both at distribution and transmission levels. Demand response aggregation and load control will enable greater grid stability and greater penetration of renewable energies into the grid. The present work fits this context in supporting charging optimization for EV in parking premises assuming a incumbent high penetration of EVs in the system. We propose a methodology to predict an estimation of the parking duration in shared parking premises with the objective of estimating the energy requirement of a specific parking lot, evaluate optimal EVs charging schedule and integrate the scheduling into a smart controller. We formalize the prediction problem as a supervised machine learning task to predict the duration of the parking event before the car leaves the slot. This predicted duration feeds the energy management system that will allocate the power over the duration reducing the overall peak electricity demand.
We structure our experiments inspired by two research questions aiming to discover the accuracy of the proposed machine learning approach and the most relevant features for the prediction models. We experiment 
different algorithms and features combination for 4 datasets from 2 different campus facilities in Italy and Brazil. Using both contextual and time of the day features, the overall results of the models shows an higher accuracy compared to a statistical analysis based on frequency, indicating a viable route for the development of accurate predictors for sharing parking premises energy management systems.
\end{abstract} 




\section{Introduction }
The advent of Electrical Vehicles (EV) are in increasing spreading in our society. According to MCkinsley report\footnote{https://www.mckinsey.com/industries/automotive-and-assembly/our-insights/mckinsey-electric-vehicle-index-europe-cushions-a-global-plunge-in-ev-sales\#} in our society EV sales rose 65 percent from 2017 to 2018 and  Europe has seen the strongest growth in EVs. According to the report "The European Union’s new emissions standard—95 grams of carbon dioxide per kilometer for passenger cars—could also boost EV sales because it stipulates that 95 percent of the fleet must meet this standard in 2020 and 100 percent in 2021".
A race for larger batteries among manufacturer is leading the current EV technology, and going forward it appears that as batteries technology improve, they are going to replace motor fuel vehicles. The concerns as we move to EVs is that, firstly, there will not be enough charge points to meet consumer demand and, secondly, this additional load on the electricity grid will cause partial and total failure of specific electrical plant due to overloading. 
One of the foreseen solutions is the use of demand response aggregation and load control \cite{LU2020117885,HERVASBLASCO2017485}. Demand response aggregators work with businesses to build a portfolio of load (or micro-generation) of which they control as per requirements of the electricity balancing market. Aggregators take payments from the the electricity market and pass these payments to their clients taking a service charge \cite{gellings2020smart}.\par
Although system services are provided to the power system via conventional power stations,  this same service is not available at distribution level. The growth in demand aggregators will enable greater grid stability and greater penetration of renewable energies into the grid \cite{mcpherson2020demand}.\par
EV energy storage capacity provides an excellent non-critical portfolio of loads which could be dynamically controlled to offer system services as well as organised in a fashion to minimise the cost of charging and reducing electricity demand peaks. Given the well documented concerns of EV penetration in the distribution system, demand response aggregators could provide an excellent control topology to employ an algorithm to safeguard and optimise the distribution network \cite{ajoulabadi2020flexible}.
The present work fits this context supporting optimization for EV charging and assuming a incumbent high penetration of EVs in the system. We propose a methodology  to predict an estimation of the parking duration in shared parking premises. This is essential for estimating the energy requirement of a specific parking lot, evaluate optimal EVs charging schedule and integrate the scheduling into a smart controller.

The specific behaviour of parking lots of campuses refereed to EV charge is peculiar since it substantially differs from the general parking lots available in the streets. In campus-like facilities (Universities, large industries, etc) we can observe regular patterns of parking behaviour that mainly include staff working hours besides a part of other visitor \cite{13561955220190401}. This can be an advantage when trying to predict  general  behavioral patterns of parking habits and thus reach an optimal recharge plan for EVs. 

The current work aims to provide an overall design of a smart charging energy management system to optimally integrate the distributed energy systems and EVs into the power grid by developing a parking prediction module to estimate the vehicles' parking time using machine learning algorithms.

Given this context, the specific objective of this work is to predict the duration of each {\em parking event} in a campus-like parking lot, where the {\em parking event} is the actual parking action of a car in a slot.
The parking duration prediction is needed to combine this information with an energy optimization algorithm that balances the energy allocation to the recharge stations with the objective for the vehicles of leaving the parking slot with a minimum charge level. While the charge optimization problem is out of the scope of this work, we now focus on the parking duration prediction. However,  we believe that the recharge optimization context is necessary to fully understand the motivation of tackling this specific prediction problem. 

We formalize the prediction problem as a supervised machine learning task that, given a parking event at a given time, tries to predict the duration of the parking event.
The reason for this event-based formulation is to be able to feed the energy management system with the duration prediction each time a car is parked. This will allow the energy management system to decide when to start the actual charge based on the prediction. For example, assuming several car arrive at 9 AM, if we start charging without a control system, the cars start charging causing a peak of electricity demand. On the contrary, if we predict that a given car stay parked e.g. 8 hours, then we can delay the charge of this car to a later moment trying to flatten the peaks.
We structure our experiments inspired by two research questions: \textbf{RQ 1}: \textit{How accurately a supervised machine learning approach can predict the duration of a parking event in a campus-like parking lot?} and \textbf{RQ 2.} \textit{What are the most relevant features for the prediction models?} 
We experiment 
different algorithms and features combination into 5 datasets from 2 different campus facilities in Italy and Brazil. We show that using both contextual and time of the day features, the overall results of the models shows an higher accuracy compared to a statistical analysis based on frequency, indicating a viable route for the development of accurate predictors for sharing parking premises energy management systems.

Structure of the paper follows. Section \ref{sec:related} discusses some related works an how this approach differentiates from them while Section \ref{sec:psystem} illustrates the design of the proposed Energy Management System for optimal charging prediction in parking premises.  Section \ref{sec:problem} introduces the problem definition and clarifies the prediction problem formulation. Section \ref{sec:experimental_evaluation} reports the details of the experimental evaluation and Section \ref{sec:discussion}. Finally, Section \ref{sec:conclusions} draws the conclusions and envisage some future works.

\section{Related Works}
\label{sec:related}

Solutions for EV optimal routes have been drafted in the logistic sectors, \cite{ZHANG2020113123,LU2020113593} accounting for charging time \cite{KARAKATIC2021114039}. However, the current work addresses a different problem that is at the crossroad of two different topics: predicting parking occupancy and predicting EV battery charging.  We report below the most relevant related works concerned these two aspects. 

\subsection{Parking availability prediction}

Most of the recent works in the literature focuses on predicting which parking lot will be free at the arrival of the car. This is motivated by the challenge of finding a parking space in urban areas. It has been reported\footnote{https://www.accessmagazine.org/spring-2011/free-parking-free-markets/} that 30\% of traffic congestion  is caused by travelling for finding parking spaces, bringing unnecessary energy consumption and environmental pollution. Works in this area include the off-street (parking slots in private areas) and in-street variants (slots in the streets). 
A pioneering paper by \cite{Caicedo12} proposes the real-time availability forecast algorithm to predict parking facility availability in real time using combined current (on-line) and historical information. This work uses an algorithm operating with mixed real and simulated information. 

In the recent paper \cite{Liao20} authors develop a  prediction model based on Naive Bayes and machine learning methods like decision tree, random forest, and regression analysis  for building the prediction model of parking occupancy and therefore predict the subsequent parking availability combining a matching-based allocation strategy to assign users to selected parking spaces.

Deep learning to predict parking occupancy is proposed by many papers in the literature. Paper \cite{Mei20} adopts a deep learning model for predicting block-level parking occupancy 30 min in advance in paper. The model takes multi-source data as input, e.g., parking, traffic and weather.


\cite{Ghosal19} proposes a Convolutional Neural Network (CNN)  model for block-level parking occupancy prediction  extracting spatial relations of traffic flow combined with a LSTM autoencoder to capture temporal correlations. Clustering is also considered in the Clustering Augmented Learning Method (CALM) to learn deep feature representations of spatio-temporal data obtained using the proposed embedding.

Paper \cite{Yuecheng18} also proposes a deep-learning approach called Du-Parking. This approach models temporal closeness, period and current general influence employing  long short-term memory (LSTM) to model the temporal closeness and period. This approach learns to dynamically aggregate the output to estimate the final parking availability of given parking lot. 


Compared to the above approaches the novelty of our method lies in the fact that we do not aim at predicting the next free slot neither to suggest the driver where to park. On the contrary, we predict how long the car will stay parked computed at the specific time a car start the parking before leaving the slot, which is a different problem. 

\subsection{EV charging profile predictions}

In \cite{CHUNG2019113732} the authors explore the best machine learning techniques for the prediction of EV charging behaviour using ensemble approach based on the data sparsity/entropy ratio. The work illustrates how such a methodology can increase the accuracy of the prediction. However, the algorithm could be difficult to scale to shared premises because it utilises only the  data from EV charging stations which represents only a small percentage of the premise.

Related works such as \cite{WU2019878} are focusing on determining the optimal charging start time assuming fixed parking time for two different group categories, day parking and overnight parking. This approach on user behaviour has been used to simplify the optimisation, but the data of a shared premise are more dynamic and so the established categories do not reflect a real scenario.

In \cite{8113485}, researchers developed an optmisation algorithm for optimal EV battery charging to reduce peak load at feeder level. Such an approach could facilitate the penetration of EV charging points in shared premises. However it does not take into account parking time prediction while it could be useful for the scheduling of energy consumption.





Compared to the above approaches our proposal is novel since we plan to embed this prediction task into an optimization algorithm of optimal charging prediction and this combination, at the best of our knowledge, this is not yet present the literature.

\section{Energy management system with prediction of the electricity demand in parking area}
\label{sec:psystem}
High penetration of EVs in shared parking lots is a likely scenario on common premises. While parking is a necessary service for offices, commercial, and residential buildings, EVs on parking site could become an asset for balancing the local power distribution network, providing ancillary services and increasing the grid's energy flexibility \cite{wevj11010018}. An Energy Management System for managing shared parking premises energy consumption will be effective only if the aggregated parking occupancy behaviour is predicted with sufficient accuracy. 
Specifically, the depicted energy management system controls the electricity demand evaluating local and power system variables, such as renewable energy production, demand response price signals and building consumption. Furthermore, it should assess the potential flexibility of the aggregated parked EVs to estimate and optimise the area's energy demand accurately. 

\begin{figure}[!ht]
	\begin{center}
	    \includegraphics[width=.98\textwidth]{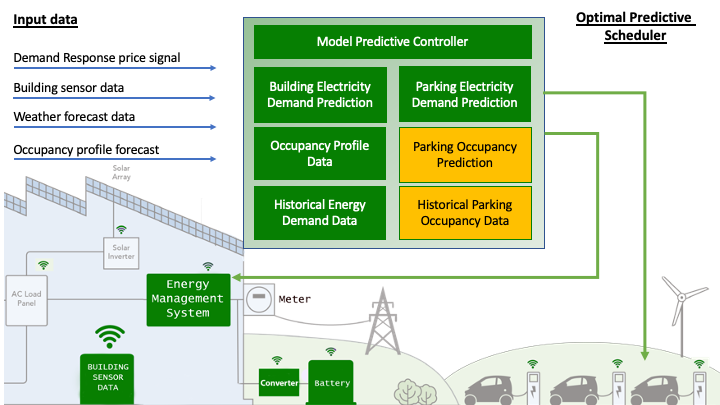}
	\end{center}
	\caption{The parking occupancy prediction module is embedded in the energy management system controller for the prediction of the electricity demand used by the parking area. In yellow the parking prediction modules proposed in the present paper}
	\label{fig:proposedsystem}
\end{figure}
Figure \ref{sec:psystem} illustrates an overall design of a distributed energy management system. The inputs data points of the system are demand response price signals, sensors data from buildings, weather forecast and images from the parking camera. The systems produces an optimal predictive schedule that balances the energy consumption while optimizing vehicle charging. 
The system objective function is to reduce the overall electricity cost and to minimise the Peak Average Ratio (PAR), defined as:
\begin{equation}
PAR={{\max_{t\in {\cal T}} \sum_{i=1}^{N}l_{i}^{t}}\over{{{1}\over{T}}\sum_{i=1}^{N}E_{i}}}
\end{equation}

where $l_{1},…,l_{N}$ are the load demand vectors for vehicle and $E_{i}$ is the predetermined electricity demand to charge the battery to the full capacity. The classical problem assumes that the vehicles will be parked on the premises long enough to be fully charged. In reality, each vehicle's parking time is variable, and it depends on several factors such as peak traffic, user habits, and weather conditions. Therefore,  the aggregated prediction of the parking time of vehicles is necessary to properly optimize the energy demand of the site. \par
In this paper we focus on a part of the energy management system dedicated to the parking occupancy prediction based on historical data, indicated in light orange in Figure \ref{fig:proposedsystem}. We present a machine learning approach for forecasting the duration of the stay of cars at the parking lot. 
The energy management system uses this information for computing the optimal changing scheduling, minimising electricity cost and PAR. The current work focuses on predicting the parking time without making any assumption on the state of charge of the battery. In a distributed energy management system, the state of charge of each vehicle is acquired as soon as the car is plugged to the charging station. Therefore, in the current work assumption on the state of charge was not critical. The objective function of the system is to maximise the energy supply to the car within the parking time reducing the overall peak demand.

\section{The parking duration prediction problem} 
\label{sec:problem}


The objective of our approach is to exploit historical data on parking usage and additional contextual data like weather conditions and parking lot occupancy levels, to predict the duration of a parking slot occupancy. 
Differently from many state of the art approaches that want to predict if a giving parking lot will be free in a next period of time \cite{Caicedo12,Liao20}, here we focus on the prediction of the temporal duration of the occupancy of a car in a slot. We recall that our approach, to be suitably integrated with an Energy Management System, focuses on specific parking context that we call of shared premises (e.g. parking lots of universities, workplaces, supermarkets, etc), not focusing on fee-based street parking.\par
Overall, our approach can be applied to any parking environment where there is a tendency for the car to stay parked a minimum amount of time and where the electrical charge system of the parking lots can be integrated into the controller of the Energy Management System. 
we also recall that our approach is driver-profile agnostic since we do not rely on driver-profile data to perform the predictions. This is because, for privacy reasons, the driver profile is generally not available. We expect that having such profile the predictions will become more accurate.

It is worth noticing that the parking behaviour in a campus-like facility reflects a different parking behaviour compared to fare-based streets parking lots. In campus-like parking, the parking duration is expected to be longer than on street parking, since  these premises are used by people parking to go to work, or study or perform an activity with a minimal temporal duration, while a fare-based parking in a street is generally affected by the parking fees that tend to encourage the reduction of the parking duration. Furthermore, the premises have usually a controlled access and the energy management system can therefore optimise the electricity supply based on the parking occupancy, while is not always true in fare-based street parking. 

\subsection{Problem formulation} 
Given  a parking area managed by an Energy Management System, a {\em car parking event} represents an event where a driver parks at a given timestamp in one of the available slots. The vehicle stays parked for a certain temporal duration until it leaves the slot. It is assumed that the vehicle can be charged while parked. The charging time can start as soon as the car arrives, or can start later on, or again, can  start, interrupt and start again. \par
Having the prediction of the parking duration when a vehicle arrives at the parking is essential to properly schedule the starting of the charge avoiding energy usage peaks.



\theoremstyle{definition}
\begin{definition}[Parking Slot] Given a parking area in shared premises, we define a parking slot $s$ as a tuple $s = <id, x, y>$, where $id$ represents the parking slot identifier, and $x$ and $y$ represent its spatial coordinates in the parking area.  
A parking slot is the actual place where drivers park their cars. In our application scenario, each slot can be equipped with a charge station where the car can be recharged. The set of all the parking slots form the parking area. 
\label{def:car_parking_slot}
\end{definition}

\theoremstyle{definition}
\begin{definition}[Car Parking Event] We define a car parking event $e$ as a tuple $e = <s_{id}, t_{start}, d_e> $, where $s_{id}$ represents the parking slot identifier where the car is parked, $t_{start}$  represents the timestamp indicating when a car has started the parking and $d_e$ is the temporal duration of the car park until it leaves the slot.
\label{def:car_parking_event}
\end{definition}

We want to predict the parking duration $d_e$ of a car parking event $e$, given a $slot_{id}$ and the parking event starting time $t_{start}$. 
This prediction is modelled as a classification problem where the objective is to assign, for each car parking event $e$, a class representing the predicted duration interval. More formally, we have the following definition of the problem. 

\theoremstyle{definition}
\label{def:problem}
\begin{definition}[Car Parking Duration Prediction Problem]  Given a parking event $e$ where it is known the slot identifier $s\_{id}$ and the start time $t_{start}$ but not duration $d_e$, we want to define a function $f(s\_{id},t_{start}) = c$ where the class $c$ represents a temporal interval such that $d_e \in c$.
\end{definition}



We can observe that our target variable $c$ represents ordinal categories. An ordinal variable is a categorical variable, where there is a clear ordering of the categories. 
For example, our variable could assume ordinal categories like: short, medium or long duration. In the next section, we introduce the details of the Machine Learning (ML) approach to solve the Car Parking Duration Prediction Problem.

\subsection{Predicting parking duration with Machine Learning} 
\label{sec:prediction}



We propose to use supervised machine learning approaches to predict the parking duration based on an historical dataset of car parking events and contextual features. 

The learning task is based on a three types of features: \emph{single event-related}, 
\emph{spatial} and \emph{contextual} features. The event-related features are detailed in Section \ref{sec:base_features} and represent the features that we can extract directly from the sets of parking events like the time of the parking event or the weather conditions.
The spatial features are based on the location of the parking slots inside the car parking area, while  the contextual features representing the occupancy of the different zones of the parking area. In Sections \ref{sec:spatial_features} and \ref{sec:contextual_features} we explain how we have extracted some spatial and contextual features that are used in our predictive models. All these features are combined to feed the proposed supervised machine learning algorithms (Section \ref{sec:experimental_evaluation}). 

\subsubsection{Extracting  event-related features}\label{sec:base_features}

Recalling Definition \ref{def:car_parking_event}, a car parking event is defined by the tuple $e = <s_{id}, t_{start}, d_e>$ and given $s_{id}$ and $t_{start}$ we want to predict the temporal duration $d_e$. 
From the timestamp $t_{start}$, we derive three features: the day of week $dw$, hour of the day $h$, and the minutes $m$ rounded to 5 minutes. The motivation of these temporal features is to enable the predictive model to learn the correlation between the time when the car parks and the relative parking temporal duration. 
We also include in this category of features the weather condition $wr$ at the moment of the car parking event starts, $t_{start}$, using this as extra information to feed the predictive models.

\subsubsection{Extracting spatial features from the parking slots locations}
\label{sec:spatial_features}

Many studies have been made toward the understating of parking behavior and the mechanism of people's parking decisions \cite{ZONG2019408, antolin2018modelling}. It has been observed that some spatial aspects can bias the occupancy of a parking lot. For example, usually drivers prefer to choose the parking areas close to the destination \cite{ma2013parking}, because they prefer to spend less time walking. However, this is not the unique criteria. Some other branch of research with focus on sustainability have investigated the impact of the trees in parking lots \cite{milovsevic2017influence}. Trees provide welcoming shade in the parking area, especially on hot days. Indeed, in the real world, there are drivers that may prefer to park the car close to the main entrance of the supermarket, while others may prefer to park in the shading areas avoiding the light of the sun during the summer.

Motivated by these aspects, our approach focus on the spatial distribution of the parking slots.
For this reason,  
we split the whole parking lot into smaller areas using different clustering approaches. Then, we include these spatial features in our predictive models to learn if
a parking area can correlate with the slot occupancy duration.

More formally, given a set of parking slots $(s_0, s_1, ..., s_n) \in S$, we use the spatial coordinates of each $s_i$
to create the spatial clusters $(a_1, a_2, .., a_k)$ of parking slots, where $k \leq n$. 
We will see in section \ref{sec:experimental_evaluation} that we have used two clustering algorithms for this task: DBScan (\cite{dbscan}) and K-Means (\cite{k-means}). Thus, when training our predictive models over the the dataset of historical car parking events 
we add as input feature a representation of the cluster where the parking slot $s_i$ belongs to. 

\begin{figure}[!ht]
	\begin{center}
	    \includegraphics[width=.98\textwidth]{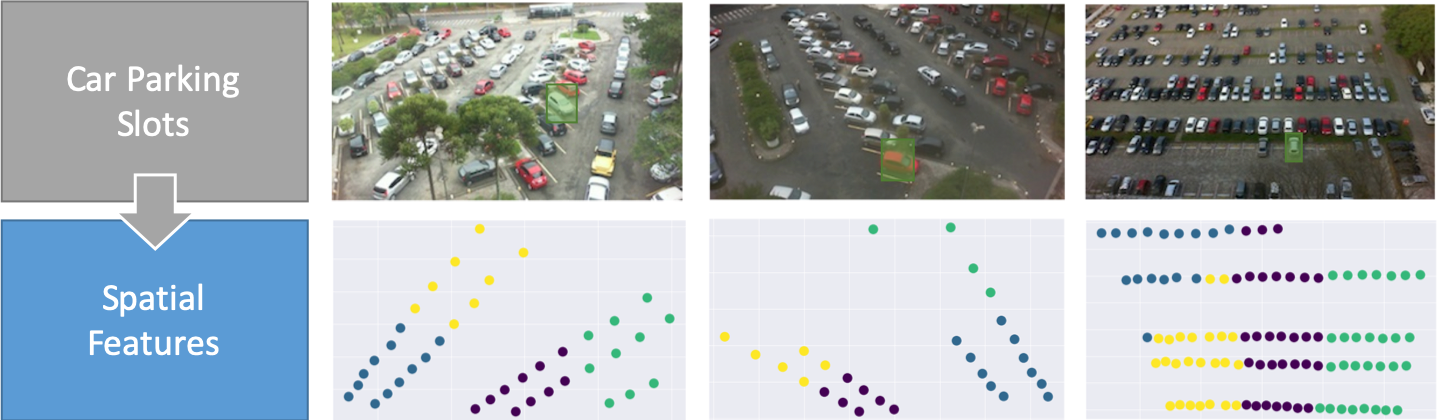}
	\end{center}
	\caption{Illustration of the spatial features using the k-means algorithm to split the parking lot into four spatial clusters.}
	\label{fig:spatial_feat}
\end{figure}

In Figure \ref{fig:spatial_feat}, we illustrate the spatial features. At the top, we can see images from three different car parking lots. At the bottom, the dots represent the pixel coordinate of the parking slots. The colors of each dots is a  representation of the spatial features which indicates the cluster (sub-area) of each parking slot. 

\subsection{Extracting contextual features }
\label{sec:contextual_features}

Another aspect that we investigate for the parking duration prediction is the context. In our case the context is represented by the status of occupancy of the slots in the spatial clusters 
and relationship of this occupancy with the duration of a given parking event. 

Specifically  we want to discover if the occupancy status of an area (e.g 100\%, means totally full, while 0\% totally empty) where a driver parks has relationship with the parking duration. 

The contextual features therefore represent the status of occupancy of the different areas (i.e. the spatial clusters) of the parking. In other words, to predict the duration of a given car parking event $e$, we also consider as input feature the level of occupancy of the spatial clusters $(a_1, a_2, .., a_k)$ at the time of the 
event. 


More formally, we define a function called $occupancy$ that given a $slot_{id}$ and the timestamp $t_{start}$ of a parking event $e$, creates a vector $(o_{a_1}, o_{a_2}, .., o_{a_k})$ by computing the occupancy level of all parking spatial clusters $(a_1, a_2, .., a_k)$ at time $t_{start}$. The occupancy status ${o_{a_j}}$ of each spatial cluster $a_j$ is basically the ratio between the number of occupied slots at the timestamp $t_{start}$ and the number of slots in that cluster.

\begin{figure}[!ht]
	\begin{center}
	    \includegraphics[width=.98\textwidth]{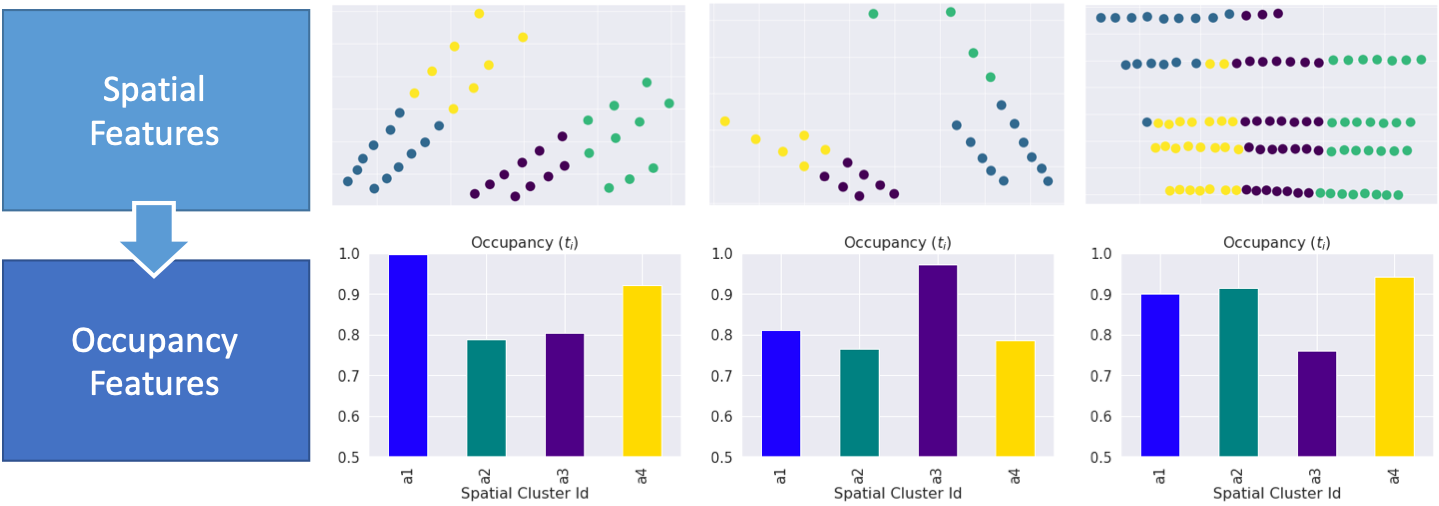}
	\end{center}
	\caption{Illustration of the contextual occupancy features representing the occupancy status of the spatial clusters $a_1, a_2, a_3,$ and $a_4$ at time $e_i.t_{start}$.}
	\label{fig:contextual_feat}
\end{figure}

In Figure \ref{fig:contextual_feat}, we illustrate the contextual occupancy features. At the top, we can see the spatial distribution of the slots from three different parking lots. The color represents the spatial cluster of each slot. At the bottom, we have the occupancy status in percent of the spatial clusters $a_1, a_2, a_3,$ and $a_4$ at a given time $t$. 

To summarize, in this section we introduced three new categories of features: event-related, spatial and contextual. The event-related features are extracted from the parking event, the spatial features are computed by using cluster techniques over the spatial distribution of the parking slots while the contextual features are obtained by computing the occupancy status of the spatial clusters. 

The overall idea is to investigate how to train the predictive models using different information that might have a predictive power on the parking duration. 
In the next section we detail the experimental setting and results on exploiting these features in a machine learning task for predicting the parking duration of a given event. 

\section{Experimental evaluation}
\label{sec:experimental_evaluation}

In this section we evaluated the proposed approach for predicting  the parking duration by exploiting historical parking data and specifically three kinds of features (see Definition \ref{def:problem}).
For a proper evaluation of the method we formulate different research questions (RQs) that we test in various scenarios by exploiting four real world and publicly available parking datasets. 

\subsection{Research Questions}
\label{sec:RQ}

The research questions driving our experiments are the following: 



\textbf{RQ 1}: \textit{How accurately a supervised machine learning approach can predict the duration of a parking event in a campus-like parking lot?}
This research question guides our first experiments. Here, we compare the performance results of our machine learning based approach against several baselines. We also investigate different machine learning approaches to tackle this problem as a supervised task: Classification, Ordinal Regression, and Regression. We use different features including the description of the occupancy of the parking lot at the time the parking event starts and the spatial distribution of the parking areas. 


\textbf{RQ 2.} \textit{What are the most relevant features for the prediction models?} We consolidate our studies investigating which are the most promising features to estimate car parking duration. We have performed a deep investigation on how the machine learning algorithms use the provided features to train the models.

\subsection{Experimental Setup}



 \textbf{Datasets.} We selected two public datasets of parking occupancy in campus-like parking lots: PKlot \cite{de2015pklot} and CNRPark \cite{Amato2016}. Both datasets contain the occupancy information detected by video cameras for each slots of parking areas of two academic institutions: the research area of the National Research Council of Pisa\footnote{http://www.area.pi.cnr.it}, in Italy and the parking area of the two Brazilian universities. In both cases the whole parking lot is split in different parking areas with a variable number of parking slots. 
In both datasets, a car parking event occurs when a car parks in a parking slot of the area. In this case, the event starts at the timestamp of the frame that detects a car in the slot. 
The car parking event ends at the timestamp of the frame showing: (1) an empty parking slot, or (2)  a different car parked in the same slot. The duration of the parking event is then computed as the difference of the timestamps of the two image frames, the start and the end.
The CNRPark dataset contains images collected from November 2015 to  February 2016 for a total of 23 monitored days. The parking lot has been monitored by 9 cameras  covering the parking area of the Pisa National Research Council south parking area. The meteorological conditions at the moment of the frame capture has also been collected. This dataset contains a total of 4081 frames and 144,965 photos. 
In Table \ref{table:CNRExcerpt} we depict an excerpt of the CNRPark raw data. 
\begin{table*}\centering
\ra{0.8}
\begin{tabular}{cccccc}\toprule
 Date & Time & Slot & Occupancy Bit & Occupancy String  & Weather\\
 \midrule
12/11/2015 & 07:15	& 275 &	0&	free &	SUNNY\\
12/11/2015 & 07:45	& 	275	& 	0 	& 	free	& 	SUNNY \\
12/11/2015 & 08:15	& 	275		& 0		& free	& 	SUNNY  \\
12/11/2015 & 08:45	& 	275		& 0		& free	& 	SUNNY \\
12/11/2015 & 09:15	& 	275		& 1		& busy	& 	SUNNY \\
12/11/2015 & 09:45	& 	275		& 1		& busy	& 	SUNNY \\
12/11/2015 & 10:15	& 	275		& 1		& busy	& 	SUNNY \\
12/11/2015 & 10:45	&   275	    & 1	    & busy	& 	SUNNY \\
12/11/2015 & 11:15	&   275	    & 1		& busy	& 	SUNNY \\
12/11/2015 & 11:45 &	275		& 1		& busy	& 	SUNNY \\
\bottomrule
\end{tabular}
\caption{An extract of the CNR parking dataset}
\label{table:CNRExcerpt}
\end{table*}


The PKlot dataset contains the occupancy information for each slot of the parking areas of two academic institutions: (1) the Federal University of Parana (UFPR) and (2) the Pontifical Catholic University of Parana (PUCPR), both located in Curitiba, Brazil. The dataset includes a total of three different parking lots represented by PUCPR, UFPR04, and UFPR05. The occupancy information is detected by a number of cameras taking images of the parking slots and detecting the change of the car or the slot becoming empty. 
 This dataset contains 12.417 images captured  in three different parking areas with different weather conditions for a total of  168 slots in the period between 11 September 2012 and 16 April 2013. Specifically, dataset PUCPR has 100 parking slots, UFPR04 has 28 and UFPR05 has 45 slots. 
PKLot is larger than CNRPark and contains images spanning across months.  


\begin{table*}\centering
\ra{0.8}
\begin{tabular}{lccccc}\toprule
Weekday &  CNRPark & PUCPR & UFPR05 & UFPR04\\
 \midrule
 Monday &	 	2  &    3 &    4 &   2\\
 Tuesday &		2 &  8 &    6 &   5 \\
 Wednesday  &			2&   5 &   4 &   4 \\
 Thursday  &				5&   5 &    4 &   4\\
 Friday  &		6&   7  &   5 & 5 \\
 Saturday  &			4&   5 & 5 & 4 \\
 Sunday &				2 &   4 &  2 & 1 \\
\bottomrule
\end{tabular}
\caption{Number of monitored days in the parking lots datasets}
\label{table:daysAllDatasets}
\end{table*}


\begin{figure}[!ht]
	\begin{center}
	    \includegraphics[width=.44\textwidth]{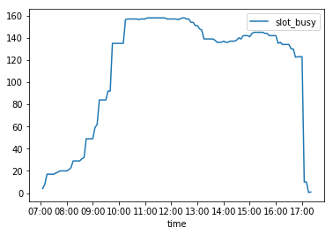}
	     \includegraphics[width=.44\textwidth]{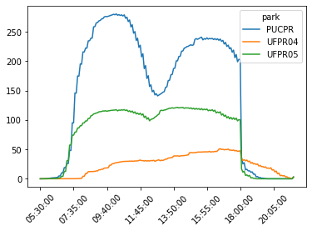}
	\end{center}
	\caption{Distribution of occupancy of the parking lots of CNR (left) and PKLot (right)}
	\label{fig:occupancydistributionCNRPKlot}
\end{figure}

\begin{figure}[!ht]
	\begin{center}
	    \includegraphics[width=.44\textwidth]{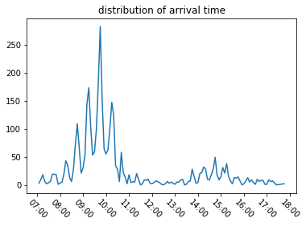}
	     \includegraphics[width=.50\textwidth]{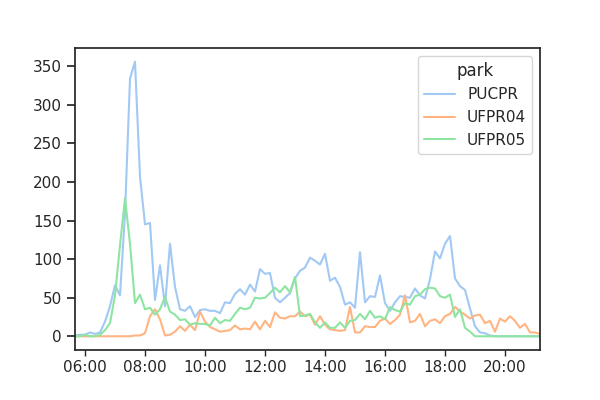}
	\end{center}
	\caption{Arrivals distribution at the parking lots of CNR (left) and PKLot (right)}
	\label{fig:arrivalsdistributionCNRPKlot}
\end{figure}

\textbf{Data Cleaning.} A detailed analysis on the image frames reveals the presence of missing data for some hour of the day (e.g. due to a broken device or during night hours due to the lack of infrared vision). Table \ref{table:daysAllDatasets} shows the total number of days of the weeks monitored in each dataset. We notice that the data collection is not uniform for each day of the week and the there is a tendency to have lesser data in the week ends.  
When a parking event is starting or ending during the missing temporal interval, it has been flagged as partial and filtered out to avoid training inconsistencies. Additionally, we filter out the days when the number of detected slots is lower than 50\% of the total slots. 
In total, after the data cleaning CNRPark counted 3552 parking events, PUCPR 4291 parking events, UFPR04 1204 parking events and  UFPR04 2148 parking events.

\textbf{Target Classes.} 
We have considered two different scenarios of classes for the predictive variable (i.e. the car parking event duration): (a) \textit{Lower sensitivity}, with longer time intervals having a total of 3 classes with discrete values in minutes:  $Short \leq 60,  60 < Mid \leq 240, Long > 240$;  and (b) \textit{Higher sensitivity} having shorter time intervals with a total of 6 classes: $Short1, \leq 30, 30 > Short2 \leq 60, 60 < Mid1 \leq 120, 120 < Mid2 \leq 240, 240 < Long1 \leq 480,$ and $Long2 > 480$. 
With these two scenarios, we want to illustrate applications requirements with different sensitivity for the predictive variable. 
Tables \ref{table:dataset_statistics_low_sensitivity} and \ref{table:dataset_statistics_high_sensitivity} show the normalized distribution of the car parking events for the low and high sensitivity scenarios respectively. Both tables also report, for each dataset, the entropy calculated according to the frequency values of the classes. We use the entropy as a measure of disorder or randomness (uncertainty) of our target variable.  


\textbf{Training Approaches.} Given the ordinal characteristic of our target variable, we have explored three supervised approaches to train the predictive models over the training set. For each approach, the best model is selected taking into account the average results over the 5 folds of validation. These approaches are: 
(a) Classification: the training is performed without taking into account the order of the classes and the selected model is the one with highest \em{micro-fscore}; (b) Regression: the training is executed to reduce the \em{ mean square error (MAE)} of the predicted values, therefore the model with lowest error is selected; 
(c) Ordinal Regression (Ord): the approach proposed in \cite{frank2001simple}  takes into account the order of the classes. The method converts the original ordinal class problem into a series of binary classification problems that encode the ordering of the original classes. This approach also aims at reducing the error and therefore selects the model with lowest {\em mean square error (MAE)}.

\begin{table*}\centering
\ra{0.8}
\resizebox{\textwidth}{!}{
\begin{tabular}{@{}lccccccccc@{}}\toprule
& \multicolumn{4}{c}{Train} & \phantom{abc}& \multicolumn{4}{c}{Test} \\
\cmidrule{2-5} \cmidrule{7-10} 
& Short & Mid & Long & Entropy  && Short & Mid & Long & Entropy \\ \midrule
CNRPark   & 0.09  &  0.31 & 0.60 & 0.81 && 0.13 & 0.36 & 0.51 & 0.89 \\
PUCPR   & 0.41  &  0.28 & 0.31 & 0.99 && 0.35 & 0.32 & 0.33 & 1.00 \\
UFPR04 & 0.52 & 0.28 & 0.20 & 0.92 && 0.44 & 0.28 & 0.28 & 0.98 \\
UFPR05  & 0.44 & 0.26 & 0.30 & 0.98 && 0.54 & 0.32 & 0.14 & 0.89 \\
\bottomrule
\end{tabular}
}
\caption{Normalized distribution of the car parking events for the low sensitivity scenario in the train and the test datasets. For each dataset we also report the entropy calculated according to the frequency values of the classes}
\label{table:dataset_statistics_low_sensitivity}
\end{table*}


\begin{table*}\centering
\ra{0.8}
\resizebox{\textwidth}{!}{
\begin{tabular}{@{}lccccccc@{}}\toprule
\cmidrule{3-8} 
 & \textbf{Short1}        & \textbf{Short2} & \textbf{Mid1}  & \textbf{Mid2}   & \textbf{Long1} & \textbf{Long2}    & \textbf{Entropy}        \\ 
&  Train/Test & Train/Test &Train/Test &Train/Test &Train/Test &Train/Test  &Train/Test\\ 
 \hline
CNRPark      & 0.06/0.10   & 0.04/0.03 & 0.09/0.10 & 0.22/0.26   & 0.53/0.57   & 0.07/0.04 & 0.76/0.78 \\
PUCPR   & 0.38/0.32    & 0.03/0.03    & 0.08/0.10    & 0.21/0.27     & 0.20/0.16   & 0.11/0.18 & 0.87/0.78                      \\
UFPR04    & 0.46/0.38  & 0.07/0.05   & 0.10/0.12      & 0.17/0.17     & 0.15/0.17     & 0.05/0.11 & 0.84/0.90      \\
UFPR05   & 0.38/0.44   & 0.06/0.10   & 0.08/0.09 & 0.18/0.22  & 0.19/0.11                                 & 0.11/0.03 &0.90/0.83                       \\ 
\bottomrule 
\end{tabular}
}
\caption{Normalized distribution of the car parking events for the high sensitivity scenario for the train and the test datasets respectively. For each dataset we also report the entropy calculated according to the frequency values of the classes}
\label{table:dataset_statistics_high_sensitivity}
\end{table*}

\textbf{Algorithms.} For the Classification and Regression tasks we used the following algorithms: Random Forest (RF), XGBoosting (XGB), AdaBoosting (AB), Logistic Regression (LR) and Support Vector Machine (SVM).  
For the Ordinal Regression task we selected: Random Forest (RF), XGBoosting (XGB), AdaBoosting (AB) and Logistic Regression (LR) 
To compute the spatial features, we have used the K-means and the DBScan clustering algorithms. For all algorithms, we used the implementation available in the scikit-learn library\footnote{https://scikit-learn.org/}. 

\textbf{Features.} 
The following features are extracted and used to feed the ML algorithms. The event-related features, as defined in Section \ref{sec:base_features}, include hour of the day $h$, time stamp minutes $m$, day of week $dw$, slot id $s$, and weather condition $wr$; the spatial features, as defined in Section \ref{sec:spatial_features}, include the spatial cluster id $spt$; the occupancy features, as defined in Section  \ref{sec:contextual_features}, include the spatial cluster occupancy $ocy$. We use different feature combinations to train the models, specifically:
\begin{enumerate}
    \item Single event-related feature. We train the model using only one event-related feature. 
    \item All event-related features together. We train the model using all single event-related features at once. We refer to $all$ when we use all the event-related features to train the ML model.
\end{enumerate}
For both cases, we perform two further combinations: using and not using the spatial and occupancy features to feed the models.

\textbf{Hyper parameters}. We use a grid search to tune the hyperparameters of the algorithms \cite{bergstra2011algorithms}. Specifically: For XGB, AB and RF, we vary the number of trees in the range of \{50, 100, 150\}, while maximum tree depth vary in the ranges of \{2,3,  or until all leaves are pure\}, respectively; For SVM, we use the RBF kernel with $\gamma$ varying in \{0.0001, 0.001, 0.01\}. For the LR, we have used two different class weight parameter \{balanced and uniform\}, while the multi class parameter changing between \{auto, ovr (for binary classification)\}. For the K-means, the $k$ varies in the ranges of \{2, 3,4,5,6\}. While for the DBScan ranges are \{50, 75, 100, 125, 150\} and \{2,3,4\} for the $\epsilon$ and minimum sample, respectively.

\textbf{Baselines.} To be able to evaluate the performance of our approach we have used the following baselines: (a) Random: randomly choose a class; (b) Longest Class: always select the longest interval; (c) Shortest Interval: always choose the shortest interval; (d) Majority Class: always choose the class with highest frequency in the training data. Furthermore, for each training approach, we also use specific baselines. For the classification and ordinal approaches, we use (e) Gaussian Naive Bayes (GNB) and (f) Multinomial Naive Bayes (MNB) as additional baseline algorithms. While for regression, we compare with the (f) Linear Regression (LN). Naive Bayes and Linear Regression are both simple ML models with high bias. They are used here as baselines given their easy interpretation.

\textbf{ML model training process.} For each dataset, 
we split the car parking events into train and test with 0.8 and 0.2 ratio respectively without shuffle the data. To avoid data leakage, we ordered the car parking events using their timestamps before split.  
When training the models on the training data, we use a stratified cross-validation with 5 folds. After the training, for each algorithm, the best configuration of hyper-parameters is used to retrain the model using the whole training data and then assess its performance now using the test set.

\textbf{Evaluation metrics}. To evaluate the experiment results we have used 
the following measures:  micro f1-score ($F1_{micro}$), macro f1-score ($F1_{macro}$) and mean absolute error (MAE). 
We recall that the F1 score (see equation \ref{lb:f1}) is a weighted average of the precision (see equation \ref{lb:precision}) and recall (see equation \ref{lb:recall}), where best value is 1 and worst is 0.
The micro f1-score is a metric where we compute an F1 score counting the total true positives, false negatives and false positives. The macro f1-score is a metric that treats all classes equally, then it does not take label imbalance into account. Indeed, the macro-average computes the metric independently for each class and then take the average, hence treating all classes equally, whereas the micro-average will aggregate the contributions of all classes to compute the average metric. 
These measures give some clues about the precision and recall of the models on predicting the true positives.
By using the MAE as in equation \ref{lb:mae} we want to have a more  interpretative
measure of our regression models  since it computes the average error of the predictions values $(\hat{y}_i)$ compared to the real values $(y_i)$. The idea is to investigate how close or how far are the predictions respect to the real values. For example, given a $Short 2$ parking event, a model that predicts it as $Mid 1$ is better than a model that predicts it as $Long 2$ therefore in this case the error is lower. 
For all experiments, we consider the MAE obtained over the test set  as comparison criteria between the models. 


\begin{equation}
    MAE = (\frac{1}{n})\sum_{i=1}^{n}\left | \hat{y_{i}} - y_{i} \right |
\label{lb:mae}
\end{equation}

\begin{equation}
    f1=2.\: \frac{precision\: .\: recall}{precision+recall}
\label{lb:f1}
\end{equation}

\begin{equation}
    precision=\: \frac{TP\:}{TP+FP}  
    \label{lb:precision}
\end{equation}

\begin{equation}
    recall=\: \frac{TP\:}{TP+FN}
\label{lb:recall}
\end{equation}



\subsection{RQ 1. Accuracy of ML in predicting parking duration}


In this section we address RQ1 - studying the accuracy of our car parking event duration prediction models. 
At this first study, 
we analyse the performance of each ML approach ($Ordinal$, $Classification,$ and $Regression$) when predicting parking events duration. Tables \ref{table:rq1_results_lower} and \ref{table:rq1_results_higher}
report the MAE, micro f1-score and macro f1-score of the models for the two different setups of classes intervals: Lower sensitivity and Higher sensitivity, respectively. The MAE was used as comparison criteria to select the best models. For each dataset and ML approach pairs, the tables indicate the strongest baseline and report the results of the best ML models for two set of features: (a) using only event-related features, represented as $Alg_{\{h,m,dw,s,wr,all\}}$, having no spatial and occupancy features; and (b) using event-related features with spatial and occupancy features, represented as $Alg^{\{spt, ocy\}}$. Both tables also report the improvement in percentage achieved by the ML models over the baselines. We highlight in bold the best MAE result per dataset.

From the results, we can observe that the ML models overcome the baselines in all the datasets, for all the training approaches. Specially the ensemble trees models (RF and XGB) show the best results  in most of the training approaches with XGB showing the best performance.
For all datasets, when using a $Regression$ approach, we observe that the most robust baseline is the Linear Regression (LN), whereas for the approaches $Classification$ and $Ordinal$ the strongest baseline is the Gaussian Naive Bayes (GNB).  
For the Lower sensitivity scenario reported in Table \ref{table:rq1_results_lower}, when predicting over the PUCPR test set, the best MAE performance (0.316) is reached by using an $Ordinal$ approach with the XGB algorithm trained with all the event-related, spatial and contextual features. This result represents an improvement of +46.3\% respect to its strongest baseline (GNB). In turn, for the Highest sensitivity scenario, as reported in Table \ref{table:rq1_results_higher}, the best MAE is 0.578 and this is the model with lowest error for the CNR dataset obtained by two ML models: $Ordinal$ with RF using all the event-related features, and $Classification$ with XGB fed with all features, including spatial and contextual. 
Moreover, we observe that the use of the spatial and occupancy features in most of the cases (16 out of 24) has improved the performance of the ML models.

When analysing the improvement of the model performance respect to the baselines, 
we observe that the highest gains were reported for the datasets with highest entropy (i.e. high uncertainty or randomness). This happened specially for the dataset PUCPR, where the ML models obtained an average gain higher than 43\%. Furthermore, in a deeper analysis, we also computed the $R^2$ of the Linear Regression model (LN) for all the datasets. We observed that the PUCPR dataset has the lowest $R^2$ values (0.004 and 0.003 for the lower and higher sensitivity scenarios respectively, against 0.35 and 0.32 computed for the CNR dataset), demonstrating the difficulty of linear models to learn the parking event duration patterns for this parking lot. Altogether, these analysis show an consistent advantage in the use of low bias ML models such as the XGB to predict parking event duration. This is due to the implicit randomness and non-linearity of such events.



\begin{table}[]
\resizebox{.99\textwidth}{!}{
\begin{tabular}{lllllll}
\hline
\textbf{Dataset}        & \textbf{Approach} & \textbf{Baseline} & \textbf{Algorithm} & \textbf{$MAE$} & \textbf{$F1_{micro}$} & \textbf{$F1_{macro}$} \\ \hline
\multirow{6}{*}{CNR}    & \multirow{2}{*}{Classification} & \multirow{2}{*}{GNB$_h$}   & $RF_h$                   & 0.345 (+5.8\%)        & 0.714 (+7.6\%)                  & 0.597 (+32.2\%)                 \\ \cline{4-7} 
                        &                                 &                          & XGB$_{all}^{ocy,spt} $       & 0.342 (+6.5\%)        & 0.714 (+7.6\%)                & 0.606 (+34.2\%)                 \\ \cline{2-7} 
  & \multirow{2}{*}{Ordinal}        & \multirow{2}{*}{GNB$_h$}   & $RF_h$                   & 0.342 (+11.6\%)*       & 0.703 (+9.4\%)                  & 0.591 (+9.6\%)                  \\ \cline{4-7} 
                        &                                 &                          & XGB$_{h}^{ocy,spt} $         & \textbf{0.336} (+13.1\%)       & 0.716 (+11.4\%)                 & 0.605 (+12.2\%)                 \\ \cline{2-7} 
                        & \multirow{2}{*}{Regression}     & \multirow{2}{*}{LN$_{all}$}  & SVM$_{all}$                & 0.417 (+11.6\%)        & 0.696 (+3.3\%)                  & 0.586 (+13.8\%)                 \\ \cline{4-7} 
                        &                                 &                          & SVM$_{all}^{ocy} $           & 0.397 (+13.2\%)         & 0.689 (+2.3\%)                  & 0.581 (+13.0\%)                 \\ \hline \hline
\multirow{6}{*}{PUCPR}  & \multirow{2}{*}{Classification} & \multirow{2}{*}{GNB$_h$}   & XGB$_{h} $                 & 0.341 (+42.0\%)       & 0.702 (+21.8\%)                 & 0.701 (+49.8\%)                 \\ \cline{4-7} 
                        &                                 &                          & XGB$_{all}^{ocy,spt} $       & \textbf{0.316} (+46.3\%)       & 0.735 (+27.5\%)                 & 0.73 (+55.9\%)                  \\ \cline{2-7} 
                        & \multirow{2}{*}{Ordinal}        & \multirow{2}{*}{GNB$_h$}   & XGB$_{all} $               & 0.336 (+42.8\%)       & 0.717 (+24.4\%)                 & 0.712 (+52.1\%)                 \\ \cline{4-7} 
                        &                                 &                          & RF$_{all}^{ocy,spt} $        & 0.317 (+46.1\%)       & 0.726 (+26.0\%)                 & 0.717 (+53.2\%)                 \\ \cline{2-7} 
                        & \multirow{2}{*}{Regression}     & \multirow{2}{*}{LN$_{all}$}  & RF$_{all}$                 & 0.39 (+43.2\%)        & 0.693 (+74.0\%)                 & 0.692 (+132.1\%)                \\ \cline{4-7} 
                        &                                 &                          & RF$_{all}^{ocy,spt} $        & 0.4 (+41.8\%)         & 0.692 (+73.7\%)                 & 0.691 (+131.6\%)                \\ \hline \hline
\multirow{6}{*}{UFPR04} & \multirow{2}{*}{Classification} & \multirow{2}{*}{GNB$_h$}   & RF$_h $                  & 0.556 (+3.6\%)*        & 0.568 (+3.0\%)                  & 0.504 (+9.8\%)                  \\ \cline{4-7} 
                       &                                 &                          & XGB$_{all}^{ocy,spt} $       & 0.51 (+11.5\%)*        & 0.593 (+7.5\%)                  & 0.547 (+19.2\%)                 \\ \cline{2-7} 
                        & \multirow{2}{*}{Ordinal}        & \multirow{2}{*}{GNB$_h$}   & XGB$_{all}$                & 0.531 (+7.9\%)        & 0.573 (+3.8\%)                  & 0.54 (+17.7\%)                  \\ \cline{4-7} 
                       &                                 &                          & XGB$_{all}^{ocy,spt}  $      & \textbf{0.506} (+12.2\%)*       & 0.556 (+0.7\%)                  & 0.525 (+14.3\%)                 \\ \cline{2-7} 
                        & \multirow{2}{*}{Regression}     & \multirow{2}{*}{LN$_{all}$}  & XGB$_{all}   $             & 0.574 (+9.7\%)        & 0.494 (+9.2\%)                  & 0.502 (+14.9\%)                 \\ \cline{4-7} 
                        &                                 &                          & XGB$_{all}^{ocy,spt}$        & 0.571 (+10.1\%)       & 0.481 (+6.4\%)                  & 0.478 (+9.2\%)                  \\ \hline \hline
\multirow{6}{*}{UFPR05} & \multirow{2}{*}{Classification} & \multirow{2}{*}{GNB$_{all}$} & RF$_h$                   & \textbf{0.340} (+16.6\%)        & 0.702 (+8.2\%)                  & 0.645 (+13.4\%)                 \\ \cline{4-7} 
                        &                                 &                          & XGB$_{h}^{ocy,spt}$          & 0.379 (+6.9\%)        & 0.672 (+3.6\%)                  & 0.6 (+5.6\%)                    \\ \cline{2-7} 
                        & \multirow{2}{*}{Ordinal}        & \multirow{2}{*}{GNB$_{all}$} & RF$_h$                   & \textbf{0.340} (+15.1\%)        & 0.702 (+7.1\%)                  & 0.645 (+11.9\%)                 \\ \cline{4-7} 
                        &                                 &                          & XGB$_{all}^{ocy,spt}$     & 0.351 (+12.2\%)       & 0.695 (+6.0\%)                  & 0.636 (+10.3\%)                 \\ \cline{2-7} 
                        & \multirow{2}{*}{Regression}     & \multirow{2}{*}{LN$_{all}$}  & RF$_h $                  & 0.463 (+11.0\%)       & 0.6 (+17.8\%)                   & 0.592 (+61.0\%)                 \\ \cline{4-7} 
                        &                                 &                          & XGB$_{all}^{ocy,spt} $       & 0.456 (+12.4\%)       & 0.593 (+16.4\%)                 & 0.571 (+55.4\%)                 \\ \hline \hline
\end{tabular}}

\caption{Car parking event duration prediction results for the lower sensitivity scenario comparing the top 1 results of each approach: Distribution, Classification, Ordinal Regression and Regression. All the improvement against the respective baseline present statistical significant differences on the residuals compared to the best baseline (Dependent t-test for paired samples’ test with 95\% confidence interval), except those indicated with *. }
\label{table:rq1_results_lower}
\end{table}

\begin{table}[]
\resizebox{.95\textwidth}{!}{
\begin{tabular}{lllllll}
\hline
Dataset        & Approach & Baseline & Algorithm & \textbf{$MAE$} & \textbf{$F1_{micro}$} & \textbf{$F1_{macro}$} \\ \hline
\multirow{6}{*}{CNR}    & \multirow{2}{*}{Classification} & \multirow{2}{*}{GNB$_h$}   & XGB$_{all} $               & 0.603 (+3.6\%)          & 0.65 (+10.3\%)                  & 0.553 (+84.8\%)                 \\ \cline{4-7} 
                        &                                 &                          & XGB$_{all}^{ocy, spt} $       & \textbf{0.578} (+7.6\%) & 0.664 (+12.7\%)                 & 0.563 (+88.3\%)                 \\ \cline{2-7} 
                       & \multirow{2}{*}{Ordinal}        & \multirow{2}{*}{GNB$_h$}   & RF$_{all} $                & \textbf{0.578} (+7.8\%) & 0.655 (+16.8\%)                 & 0.468 (+56.8\%)                 \\ \cline{4-7} 
                        &                                 &                          & RF$_{all}^{ocy, spt}$         & 0.580 (+7.6\%)          & 0.668 (+19.0\%)                 & 0.567 (+89.7\%)                 \\ \cline{2-7} 
                        & \multirow{2}{*}{Regression}     & \multirow{2}{*}{LN$_{all}$}  & SVM$_{h} $               & 0.648 (+12.8\%)          & 0.594 (+27.9\%)                 & 0.365 (+42.2\%)                 \\ \cline{4-7} 
                        &                                 &                          & SVM$_{h}^{ocy}$        & 0.654 (+12.0\%)          & 0.588 (+26.7\%)                 & 0.361 (+40.7\%)                 \\ \hline \hline
\multirow{6}{*}{PUCPR}  & \multirow{2}{*}{Classification} & \multirow{2}{*}{GNB$_h$}   & XGB$_{all}$                & 0.703 (+48.9\%)         & 0.64 (+53.6\%)                  & 0.44 (+145.5\%)                 \\ \cline{4-7} 
                        &                                 &                          & XGB$_{all}^{ocy, spt}  $      & \textbf{0.672} (+51.2\%)         & 0.65 (+55.9\%)                  & 0.437 (+143.8\%)                \\ \cline{2-7} 
                        & \multirow{2}{*}{Ordinal}        & \multirow{2}{*}{GNB$_h$}   & RF$_h$                   & 0.723 (+44.3\%)         & 0.633 (+18.8\%)                 & 0.421 (+80.9\%)                 \\ \cline{4-7} 
                        &                                 &                          & XGB$_{all}^{ocy, spt} $       & 0.693 (+46.6\%)         & 0.657 (+23.1\%)                 & 0.452 (+94.0\%)                 \\ \cline{2-7} 
                        & \multirow{2}{*}{Regression}     & \multirow{2}{*}{LN$_{all}$}  & RF$_{all} $                & 0.781 (+48.7\%)         & 0.511 (+109.0\%)                & 0.431 (+208.6\%)                \\ \cline{4-7} 
                        &                                 &                          & XGB$_{all}^{ocy, spt} $       & 0.833 (+45.2\%)         & 0.473 (+93.3\%)                 & 0.404 (+189.8\%)                \\ \hline \hline
\multirow{6}{*}{UFPR04} & \multirow{2}{*}{Classification} & \multirow{2}{*}{GNB$_{all}$} & RF$_h$                   & 1.348 (+11.0\%)         & 0.423 (+3.0\%)                  & 0.178 (+-8.6\%)                 \\ \cline{4-7} 
                        &                                 &                          & XGB$_{all}^{ocy, spt} $       & \textbf{1.166} (+23.0\%)         & 0.486 (+18.2\%)                 & 0.292 (+50.2\%)                 \\ \cline{2-7} 
                        & \multirow{2}{*}{Ordinal}        & \multirow{2}{*}{GNB$_h$}   & XGB$_{all}$                & 1.249 (+9.3\%)          & 0.465 (+4.7\%)                  & 0.232 (+17.5\%)                 \\ \cline{4-7} 
                        &                                 &                          & XGB$_{all}^{ocy, spt}  $      & 1.178 (+14.5\%)         & 0.461 (+3.7\%)                  & 0.283 (+43.4\%)                 \\ \cline{2-7} 
         & \multirow{2}{*}{Regression}     & \multirow{2}{*}{LN$_{all}$}  & XGB$_{all}$               & 1.252 (+9.3\%)          & 0.191 (+53.3\%)                 & 0.167 (+42.0\%)                 \\ \cline{4-7} 
                        &                                 &                          & XGB$_h^{ocy, spt}   $       & 1.246 (+9.7\%)          & 0.183 (+46.7\%)                 & 0.186 (+58.0\%)                 \\ \hline \hline
\multirow{6}{*}{UFPR05} & \multirow{2}{*}{Classification} & \multirow{2}{*}{GNB$_{all}$} & RF$_h$                   & \textbf{0.854} (+23.4\%)         & 0.567 (+25.1\%)                 & 0.322 (+45.4\%)                 \\ \cline{4-7} 
                        &                                 &                          & XGB$_h^{ocy, spt}  $        & 0.977 (+12.3\%)         & 0.535 (+17.9\%)                 & 0.313 (+41.4\%)                 \\ \cline{2-7} 
                        & \multirow{2}{*}{Ordinal}        & \multirow{2}{*}{GNB$_{all}$} & RF$_h$                   & \textbf{0.854} (+16.6\%)         & 0.567 (+8.0\%)                  & 0.322 (+29.7\%)                 \\ \cline{4-7} 
                        &                                 &                          & XGB$_h^{ocy, spt} $         & 0.93 (+9.1\%)           & 0.558 (+6.2\%)                  & 0.322 (+29.8\%)                 \\ \cline{2-7} 
                        & \multirow{2}{*}{Regression}     & \multirow{2}{*}{LN$_{all}$}  & XGB$_{all} $               & 0.987 (+13.3\%)         & 0.323 (+40.4\%)                 & 0.232 (+41.9\%)                 \\ \cline{4-7} 
                        &                                 &                          & XGB$_h^{ocy, spt}  $        & 0.975 (+14.4\%)         & 0.346 (+50.5\%)                 & 0.284 (+74.1\%)                 \\ \hline
\end{tabular}}
\caption{Car parking event duration prediction results for the higher sensitivity scenario. Comparing the top-1 results of each approach: Distribution, Classification, Ordinal Regression and Regression. All the improvements against the respective baseline present statistical significant differences on the residuals compared to the best baseline (Dependent t-test for paired samples’ test with 95\% confidence interval).} 
\label{table:rq1_results_higher}
\end{table}

\subsection{RQ 2. Measuring Features importance}


In section,  addressing Research Question N. 2, we study the  feature importance scores of the best models for each dataset and sensitivity scenario.  
Feature importance scores can provide insights of how the ML models works and what can be further improved. The relative scores can highlight which features are most relevant for the model to predict the target values, and the converse, which features are the least relevant.



Tables \ref{tab:rq3} \ and \ref{tab:rq3-2} show the relative importance score of  the best predictor of each dataset in the two scenarios. These measures allows us to discover what are the most predictive features in our datasets. The higher is the feature's importance, the more informative the feature is for the prediction of the parking event duration.
In most of the datasets, the hour of the day $h$ is the most important feature, indicating a relevant relationship between the time that the parking event happens and its duration. Indeed, in the UFPR05 dataset, for both the lower and higher sensitivity scenarios, the best model was achieved by only using the hour of the day $h$ as a feature. The day of the week $dw$ has shown low relevance having importance values close to zero. This means that no relevant changes concerning how the parking events happen along the week's days have been observed in our datasets. 
When analysing the features with spatial properties $s$ indicating the parking slot and the spatial cluster $spt$, we can observe that the ML models cannot learn individual slot patterns in general. Indeed, the feature $s$, that discriminates the single parking slot, has shown very low relevance having most of the values close to zero. This means that the models did not find distinguishable patterns over individual slots. 
These results could be potentially different in the case of parking lots where the parking slots are exclusively assigned to specific drivers. 
Differently, for the experiments considering the spatial cluster $spt$, the ML model assigns a relevant weight to this feature. This means that the models were able to learn and use the dynamic of the parking events within the group of slots (i.e. areas of the parking lot) to predict the duration of parking events.


\begin{table}[]
\resizebox{.99\textwidth}{!}{
\begin{tabular}{llllllllll}\toprule
\textbf{Dataset} & \textbf{Approach} & \textbf{Algorithm} & \textbf{h}     & \textbf{tm} & \textbf{dw} & \textbf{s} & \textbf{wr} & \textbf{spt} & \textbf{ocy} \\ \midrule
CNRPark              & Ordinal           & XGB$_{h}^{ocy,spt}  $              & \textbf{0.547} & 0.000       & 0.000       & 0.000      & 0.000       & 0.230            & 0.223              \\ \hline
PUCPR            & Classification    & XGB$_{all}^{ocy,spt}  $              & \textbf{0.467} & 0.045       & 0.072       & 0.041      & 0.098       & 0.077            & 0.200              \\ \hline
UFPR04           & Ordinal           & XGB$_{all}^{ocy,spt}  $              & 0.181          & 0.059       & 0.063       & 0.055      & 0.242       & 0.110            & \textbf{0.290}     \\ \hline
UFPR05           & Ordinal           & RF$_{h} $                & \textbf{1.000} & 0.000       & 0.000       & 0.000      & 0.000       & 0.000            & 0.000              \\  \hline
\end{tabular}}
\caption{Relative importance of the variables for the lower sensitivity scenario}
\label{tab:rq3}
\end{table}

\begin{table}[]
\resizebox{.99\textwidth}{!}{
\begin{tabular}{llllllllll}\toprule
\textbf{Dataset} & \textbf{Approach} & \textbf{Algorithm} & \textbf{h}     & \textbf{tm} & \textbf{dw} & \textbf{s} & \textbf{wr} & \textbf{spt} & \textbf{ocy} \\ \midrule  
CNRPark              & Classification    & XGB$_{all}^{ocy,spt} $               & \textbf{0.410} & 0.064       & 0.054       & 0.036      & 0.081       & 0.186            & 0.170              \\ \hline
PUCPR            & Classification    & XGB$_{all}^{ocy,spt} $               & 0.284 & 0.031       & 0.038       & 0.027      & 0.077       & 0.132            & \textbf{0.410}              \\ \hline
UFPR04           & Classification    & XGB$_{all}^{ocy,spt}   $             & 0.157          & 0.059       & 0.049       & 0.043      & 0.231       & 0.166            & \textbf{0.294}     \\ \hline
UFPR05           & Classification    & RF$_{h}  $               & \textbf{1.000} & 0.000       & 0.000       & 0.000      & 0.000       & 0.000            & 0.000              \\ \hline
\end{tabular}}
\caption{Relative importance of the variables in the higher sensitivity scenario}
\label{tab:rq3-2}
\end{table}

\begin{figure}[!ht]
	\begin{center}
	    \includegraphics[width=.98\textwidth]{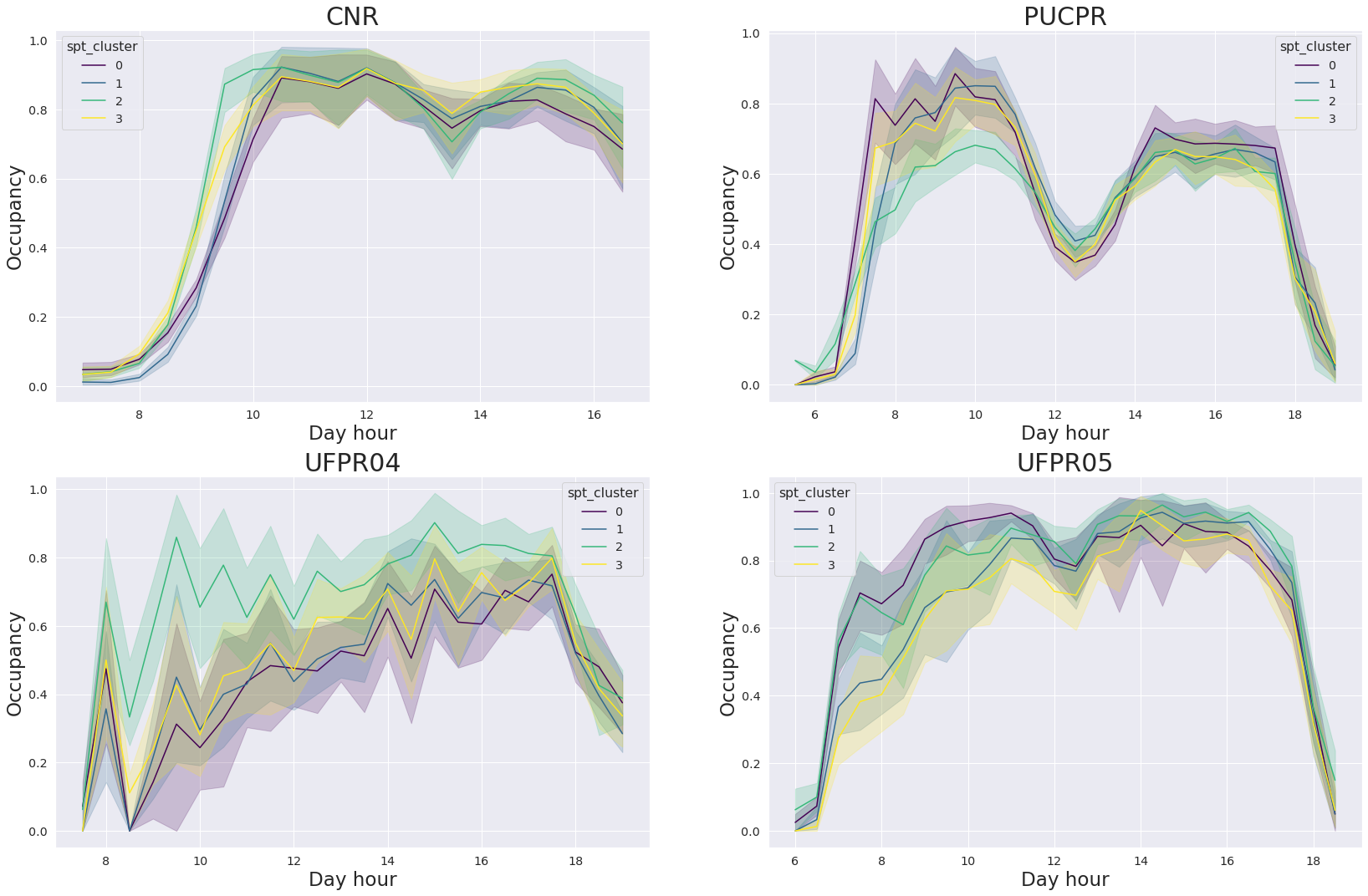}
	\end{center}
	\caption{Visualization of the occupancy features values over the time of the day of four spatial clusters defined by using the Kmeans algorithm. The plot help us to show the non-perfect linear relationship between the time and the occupancy of the parking. }
	\label{fig:occupancy_feat}
\end{figure}

Overall, for all the datasets, the occupancy feature $ocy$ has shown significant importance, having the highest straight for the PUCPR dataset in the higher sensitivity scenario. However, it has to be noted that the occupancy feature $ocy$ is computed from the parking event starting time, therefore, a further study is needed to to analyse the correlation between $ocy$ and $h$.  
In regression analysis, we use the term {\em multicollinearity} to indicate a model including multiple features that are correlated, not to the target variable, but also to each other. The plot in Figure \ref{fig:occupancy_feat}  helps us in this analysis. The plot shows, for each dataset, the occupancy status in percentage for four different spatial clusters defined by the algorithm K-means. We can observe that there is no strict correlation between the hour of the day and the occupancy of the parking lot areas.  
We see that the occupancy increases from beginning of the day, then it drops at lunchtime and then again late in the afternoon. These drops in occupancy over the day clearly show a non-redundant relationship between the $h$ and $ocy$, evidencing then that their effects on the ML models are actually distinguishable.




\section{Discussions}
\label{sec:discussion}

\begin{figure}[!ht]
	\begin{center}
	    \includegraphics[width=.98\textwidth]{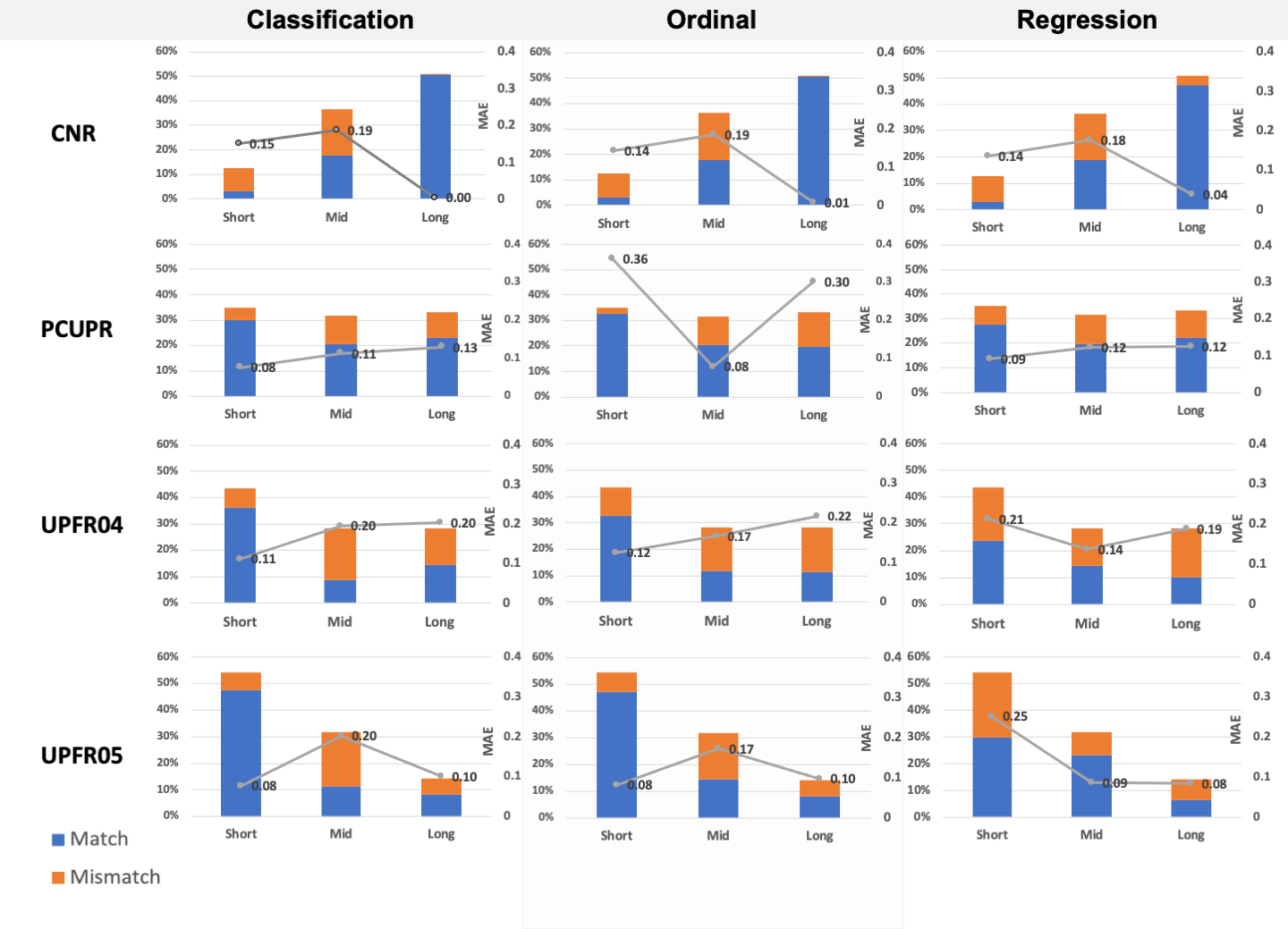}
	\end{center}
	\caption{For lower sensitivity (3 intervals) and for the four datasets analysed, comparison of the accuracy and MAE for the prediction}
	\label{fig:3cat}
\end{figure}

The use of EV batteries as flexible energy storage opens up several research questions on integrating charging vehicles with the electrification of buildings and users' transport patterns. At the core of these challenges there is the development of smart energy management systems. An energy management system (EMS) monitors, controls, and optimises the distributed resources' operation to achieve specific operational and economic objectives (e.g., minimise costs).  As discussed in Section \ref{sec:psystem}, the proposed energy management system facilitates EVs' integration as movable V2G assets. Such a system requires implementing a parking occupancy prediction module based on historical data, identifying driver behaviours, and using machine learning algorithms for forecasting the parking time for each parking slot. The analysis of various machine learning algorithms across four different parking datasets to predict parking behaviours allowed us to establish a methodology to develop the energy management system's parking prediction module. Figure \ref{fig:3cat} shows the accuracy and the mean average error of the three 
machine learning approaches tested against the four datasets covering the lower sensitivity categories (3 categories). For this analysis, we use the prediction values of the best models reported in Tables \ref{table:rq1_results_lower} and \ref{table:rq1_results_higher}. In such a case, the identification of the best prediction algorithm is not straightforward. The compared results highlight each dataset's limitation, providing an insight into the data quality and consistency of behaviour. The events of the CNR dataset result in the most difficult to predict, especially in the short term parking category. On the opposite, UPFR05 looks more predictable among the four.\par
Both classification and regression algorithms produce a similar performance for long term parking; however, classification is more accurate on the short term forecasting while regression has an overall lower mean average error in the medium range. The parking prediction module based on classification could provide a better user experience to drivers because accurate identification of short-term parking will force the controller to guarantee a higher energy share to short-term park events. However, it will reduce the peak shaving capabilities of the parking area. On the other hand, a regression model could facilitate demand response measures because of forecasted parking events shifted towards long parking time.\par
\begin{figure}[!ht]
	\begin{center}
	    \includegraphics[width=.98\textwidth]{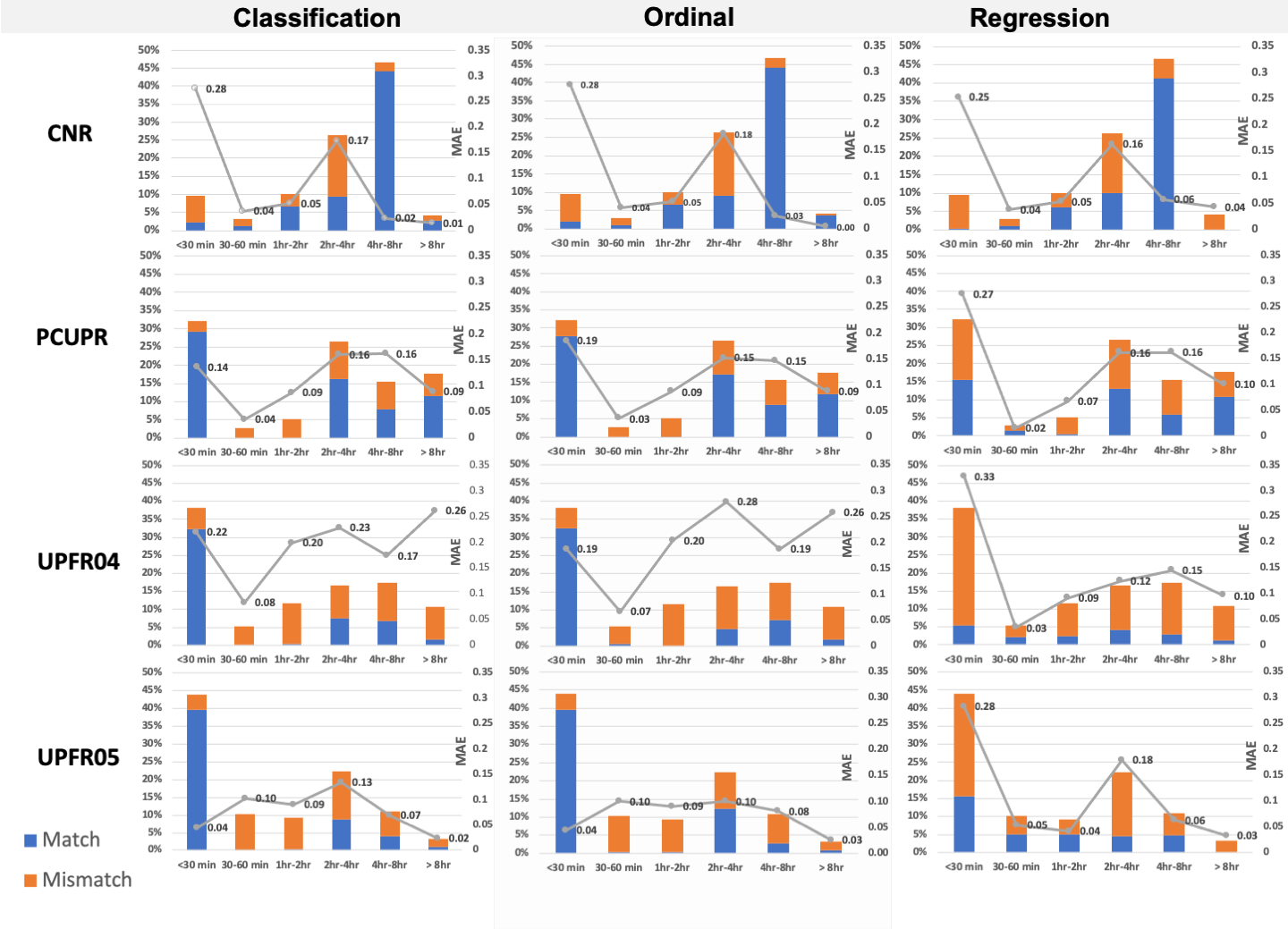}
	\end{center}
	\caption{For higher sensitivity (6 intervals) and for the four datasets analysed, comparison of the accuracy and MAE for the prediction }
	\label{fig:6cat}
\end{figure}

Figure \ref{fig:6cat} shows the predictive algorithms' results on the higher sensitivity categories (6 categories). The results confirm the challenges in forecasting short term parking events for the CNR dataset while it confirms the best prediction performance for the UPFR05 data. In this context, both classification and ordinal approaches are the most accurate for short term parking events ($<$30 mins), while regression is the most accurate for long term parking events ($>$ 8hr). In the latter case, the ordinal approach is the more accurate for three datasets over four, while predicting UPFR04 shows a high percentage of errors. All the approaches result in low accuracy for categories with a smaller number of events, such as categories 2 (30-60 min) and 3 (1hr-2hr). Such a lower score depends on the limited amount of data available for training. In comparison, high-frequency events are most likely to be correctly predicted, as illustrated in the CNR dataset, category 5 (4-8 hrs) and in UPFR04, UPFR05 and PCUPR for category 1 ($<$30 min). Overall, the lower sensitivity models are more accurate that to be integrated in an optimisation module for energy management systems. The high sensitivity predictions result suffering from low accuracy and high mean average error that could lead to system malfunctions and uncertainty. Therefore, to integrate the parking prediction module in an energy management system, further improvement of the forecasts are necessary, especially for high sensitivity experiments. Additional data could be used to improve the prediction's accuracy, such as higher picture resolution that could better identify users by reading the licence plates or identify unique marks (stickers, internal objects or scratches). Additionally, other data sources can be employed to forecast the number of car at an aggregated level and compare it with similar works.
The current work aims to provide an overall design of a smart charging energy management system to optimally integrate the distributed energy systems and EVs into the power grid by developing a parking prediction module to estimate the vehicles' parking time using machine learning algorithms. The proposed system can capture EVs users' aggregated uncertain behaviour to obtain an optimised solution for both the capital expenditures (CAPEX) and operational expenditures (OPEX) at the network's planning and operation phases. CAPEX can be minimised by optimising and distributing distributed energy resources and charging stations for electric vehicles under economic and social constraints. At the same time, optimal power flow solutions considering technical constraints can lead to OPEX minimisation.

\section{Conclusions and Future Works}
\label{sec:conclusions}
The provision of energy system services plays a critical role for the power system as renewable energy sources are displacing conventional generation, and system demand is soaring due to the electrification of transport and thermal loads. The provision of energy services using demand response programs is an area of great potential for system operators; however, there remain several challenges in proving the effectiveness of such schemes integrated with the transport systems. The growing penetration of EVs with larger battery size challenges the distribution network's capacity, and it is becoming a threat to the grid's reliability. Nevertheless, suppose electric vehicles enable V2G operations, and parking premises provide a data harvesting infrastructure to produce accurate parking slot predictions. In that case, such a mobile storage capacity can become an asset to improve grid reliability and resiliency and used to defer capital upgrade costs.\par
Future works include the use of anonymised user profiles to reach more accurate predictions based on the single user habits, as well as having more dense and richer datasets to improve the accuracy of the models. Another direction is the proper integration of the prediction module into an Energy Management System as illustrated above in Section \ref{sec:psystem}. Specifically, when a car starts its parking at a parking slot (detected by a camera or by the connection to the charging station) the prediction of the parking duration will feed, together with the energy models and other contextual data, the model predictive controller that will allocated the energy to the charge station in the slots. 
This will foster new research directions like for example a combined model of parking duration prediction and energy optimization models to solve the power supply-demand balancing problem by aggregating EVs, renewable energy systems, energy storage, and demand response at distribution level. Additionally, the energy management system equipped with the module for the parking prediction can trade the expected energy consumption in the electricity market reducing the overall energy cost.
 
\section*{Acknowledgment}
The work is supported by the ERA-NET Smart Energy System, Sustainable Energy Authority Ireland and Italian Ministry of Research with project N. ENSGPLUSREGSYS18\_00013. This publication has emanated from research conducted with the financial support of of the EVCHIP project under grant agreement 19/RDD/579, EVCHIP.

\end{document}